Original Research Paper

# A Novel Dialect-Aware Framework for the Classification of Arabic Dialects and Emotions

**Nasser Alsadhan**

*Department of Computer Science, King Saud University, Riyadh, Saudi Arabia*

**Abstract:** Arabic is one of the oldest languages still in use today. As a result, several Arabic-speaking regions have developed dialects that are unique to them. Dialect and emotion recognition have various uses in Arabic text analysis, such as determining an online customer's origin based on their comments. Furthermore, intelligent chatbots that are aware of a user's emotions can respond appropriately to the user. Current research in emotion detection in the Arabic language lacks awareness of how emotions are exhibited in different dialects, which motivates the work found in this study. This research addresses the problems of dialect and emotion classification in Arabic. Specifically, this is achieved by building a novel framework that can identify and predict Arabic dialects and emotions from a given text. The framework consists of three modules: A text-preprocessing module, a classification module, and a clustering module with the novel capability of building new dialect-aware emotion lexicons. The proposed framework generated a new emotional lexicon for different dialects. It achieved an accuracy of 88.9% in classifying Arabic dialects, which outperforms the state-of-the-art results by 6.45 percentage points. Furthermore, the framework achieved 89.1-79% accuracy in detecting emotions in the Egyptian and Gulf dialects, respectively.

## Introduction

As languages develop across regions far apart, dialects begin to take shape. A dialect is a variation of a language in grammar, pronunciation, and vocabulary. Every individual has a way of talking that reflects their dialect, accent, background, and many other factors Biadsy (2011). The Arabic language exhibits a variety of dialects throughout the Arab world; dialects can differ not only across countries but also within the same country or city. In particular, Arabic dialects differ from one another in pronunciation and vocabulary. Therefore, different dialects have different words and variations of words that could refer to the same meaning. These discrepancies make it difficult for people with different dialects to understand each other. It is even harder for non-Arabic speakers who recently learned Arabic. Dialects developed mainly after the foundation of independent Arab countries. This separation has reduced the interaction between different regions and as a result, many Arabic-speaking regions have developed dialects exclusive to their own. For example, Arabic evolved in many countries surrounding the Arabian Gulf, yielding dialects different from those used in the Levantine region.

One of the main problems in understanding the emotions and meaning behind Arabic text using machine learning is the variety of dialects and sub-dialects in Arabic. Natural Language Processing (NLP) has been extensively used to solve such problems. Most of the work done in this area of research is applied to the English language. The Arabic language and its different dialects remain an open area of research regarding detecting dialects, emotions, and complex concepts such as irony and sarcasm. Researchers in machine learning have become interested in such problems for the benefits they can provide. For example, by identifying the region where customers come from, companies can analyze a product's reviews and comments and break them down by region, which provides valuable intel for a business. It also helps in predicting the nationality of an anonymous writer of a piece of text by predicting their region. In addition, understanding the emotions and meaning of a text can help us build better chatbots that are aware of someone's emotions, which enables the chatbot to respond appropriately. Furthermore, such analysis can be applied to detect the overall emotional tone of the public in different areas of the Arab world.

The main two problems addressed in this study are: (i) How to identify and predict Arabic dialects from a piece of text; and (ii) How to identify emotions in different dialects. Mainly, the paper addresses those challenges by building a novel framework that analyzes and understands Arabic text and classifies its main dialects and the emotions conveyed by different dialects. The main contribution of this study is the ability to generate a dialect-aware emotional lexicon from any given corpus labeled with the dialects it exhibits.

*Literature Review*

Dividing Arabic into different dialects to distinguish them from Modern Standard Arabic (MSA) is a complex task, as dialects shift and change depending on the time and precision intended to be administered in the breakdown. Researchers working on this problem have found various breakdowns. Habash (2010) has suggested the following breakdown while noting that it should not be assumed that all members of any dialect group are linguistically identical:

- Egyptian Arabic (EGY) refers to Egypt and Sudan
- Gulf Arabic (GLF) refers to the Arabic peninsula
- Levantine Arabic (LEV) refers to the Levantine region
- North African (MAGHREBI) refers to Morocco, Algeria, Tunisia, and Mauritania
- Iraqi Arabic (IRQ) has elements of both Levantine and Gulf
- Yemenite Arabic (YEM) is often considered its own class

Zaidan and Callison-Burch (2011) gave a similar breakdown to Habash's, which is given in Fig. (1). Another breakdown was given by Alshutayri (2018) which includes GLF (including Oman), EGY, LEV, NOR (which includes Morocco, Algeria, Tunisia and Libya) and IRQ. Alsudais *et al*. (2022) investigated the geographical similarity of various Arabic dialects. They concluded that the correlation between dialectical similarity and city proximity suggests that cities close to each other are more likely to share dialectical characteristics, regardless of their origin. Although these breakdowns are somewhat general and imprecise, they are general enough to be helpful in data collection and classification.

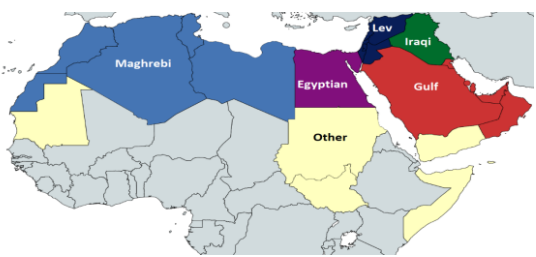

**Fig. 1:** Geographical region for each dialect using Zaidan's breakdown

The problem of dialect classification has been studied in the past, with many studies building their corpora. Zaidan and Callison-Burch (2011) collected the most prominent corpora, the Arabic Online Commentary (AOC) dataset, which gathered millions of comments from three newspapers. Though the AOC dataset was substantial, it needed to be fully annotated. Further work to annotate AOC was done by El-Haj *et al*. (2018). They used Amazon's Mechanical Turk (MTURK), which hires online annotators. Another improvement in the AOC dataset came from Cotterell and Callison-Burch (2014). They extended the AOC newspaper dataset to include five newspapers: An Egyptian newspaper "Al-Youm Al-Sabe"; a Saudi-Arabian newspaper "Al-Riyadh"; a Jordanian newspaper "Al-Ghad"; an Algerian newspaper, "Each Chorouk El Youmi"; and an Iraqi newspaper "Al-Wefaq"; as well as tweets scraped from Twitter. After collecting the extended dataset, they annotated them using MTURK. Some research focused on using social media as a valuable source of dialectic data. One of the datasets is the Social Media Arabic Dialect Corpus (SMADC), which scraped and annotated data from Twitter and Facebook Alshutayri and Atwell (2019). Another social media dataset is the Dialectal Arabic Tweets (DART), which manually annotated over 25k tweets in Maghrebi, Egyptian, Levantine, Iraqi, and Gulf Alsarsour *et al*. (2018).

Zaidan and Callison-Burch (2011) used the Stanford Research Institute Language Modeling (SRILM) toolkit on the AOC dataset to predict Arabic dialects. They built word trigram models with modified Kneser-Ney as a smoothing method. They reported the results of 10-fold cross-validation, which achieved an accuracy of 79.6% for MSA vs. LEV, 75.1% for MSA vs. GLF, and 80.9% for MSA vs. EGY. Cotterell and Callison-Burch (2014) used the same dataset above and experimented with two algorithms, Support-Vector Machine (SVM) and Naive Bayes (NB), using unigram, bigram, and trigram features. The NB method using uni-gram outperformed the other approaches. Their results also outperformed Zaidan and Callison-Burch (2011), with an accuracy of 86.6% for MSA vs. LEV, 82.7% for MSA vs. GLF, and 86.6% for MSA vs. EGY. The major drawback of these studies is that multiple classifiers exist for the same problem and they can only distinguish between MSA and one other class simultaneously.

Alshutayri (2018) used the SMADC dataset to classify dialects into GLF, NOR, LEV, EGY, and IRQ. They used the Sequential Minimal Optimization (SMO) algorithm with multinomial Naive Bayes (MNB) with different tokenizers. They ran their analysis via Waikato Environment for Knowledge Analysis (WEKA) to achieve an accuracy of 60.7%. They improved their results with a lexical method to achieve an accuracy of 69.2%. It consisted of a simple voting mechanism, which counts the number of words belonging to a given dialect. An example of simple voting is shown in Table (1).

**Table 1:** Simple voting matrix representation

| Words | NOR | EGY | IRQ | LEV | GLF |
|---|---|---|---|---|---|
| ههههههه | 0 | 1 | 1 | 1 | 1 |
| وقت | 0 | 0 | 0 | 0 | 0 |
| ليش | 1 | 0 | 1 | 1 | 1 |
| بالهطريقة | 0 | 0 | 1 | 0 | 1 |
| صافي | 1 | 0 | 0 | 0 | 0 |
| ده | 0 | 1 | 0 | 0 | 0 |
| زاكي | 0 | 0 | 0 | 1 | 0 |
| Total | 2 | 2 | 3 | 3 | 3 |

Most of the research around dialect classification uses traditional text classification methods, while deep learning methods are scarcely used. However, the surge of deep learning research has reinvigorated the interest in using it in Arabic dialect classification. Elaraby and Abdul-Mageed (2018); and Lulu and Elnagar (2018) have used deep learning algorithms and traditional classifiers such as SVMs and NB. They used the AOC dataset with the model predicting one of four classes (MSA, EGY, GLF, and LEV). The best-performing model was obtained by Elaraby and Abdul-Mageed (2018) using a Bidirectional Long Short-Term Memory (BiLSTM), with an accuracy of 82.45%.

Work-related to detecting the psychological features of the author of an Arabic piece of text is relatively new Alsadhan and Skillicorn (2017); Sharmila *et al.* (2019). The two main problems faced in this research domain are (i) The need for more adequately annotated data, and (ii) The role different Arabic dialects play in emotional and psychological tone. Most of the work related to emotion detection in Arabic ignores the type of dialect used. Al-Khatib and Al-Beltagy (2018) collected and manually annotated 10k Arabic tweets with six emotions (anger, disgust, fear, happiness, sadness, and surprise). They achieved an accuracy of 68.1% using NB as their algorithm. Eslam *et al.* (2019) used transfer learning and Convolutional Neural Networks (CNN) to predict the emotion of a piece of text. They used the data collected by Al-Khatib and El-Beltagy (2018). Their approach achieved a high accuracy of 95.2%, but due to the small size of the dataset (10k tweets) and the complexity of their model, their results might not generalize well. Rabab'ah *et al.* (2016) annotated another dataset comprising 2025 tweets and four emotions (Sadness, joy, disgust, and anger). This was used by Abdullah *et al.* (2020) using different traditional machine learning models, and the best result was reported using SVM with an accuracy of 80.6%. The most comprehensive Arabic Emotion Lexicon (AEL) was created by Saad (2015). It contains 3207 words and six emotions. It was built by translating Ekman's basic human emotions (anger, disgust, fear, joy, sadness, and surprise) Ekman (1992). The two main drawbacks of the studies mentioned above are the need for a substantial amount of annotated data and dialect awareness.

A few studies have examined the use of dialects when detecting emotions. Such as the work done by Dahmani *et al.* (2019), which detects emotions in the Algerian dialect using audio features. Their data consists of TV show recordings and five emotions (admiration, disapproval, enthusiasm, joy, and neutral). Aljuhani *et al.* (2021) collected YouTube videos in a Saudi dialect and manually annotated them with four emotions (anger, happiness, sadness, and neutral). They used audio features and studied the effectiveness of different models. Their best model was SVM, with an accuracy of 77.1%. Other work focuses on building publicly available data for Arabic dialects and emotions. Moudjari *et al.* (2020) proposed a collaborative annotation platform for crowdsourcing annotation called TWItter proFILing (TWIFIL). Their initial work created a considerable annotated corpus for the Algerian dialect. They aim to include more NOR dialects into their platform in the future. TWIFIL can be exploited for opinion and emotion analysis at a relatively low cost and can be used as a benchmark dataset when enough dialects are collected within it.

*Proposed Approach*

In order to address the limitations of the state-of-the-art depicted in Section 2, the paper proposes a novel framework able to identify and predict Arabic dialects and emotions conveyed by a given text. Precisely, the proposed framework consists of three components: (i) A text-preprocessing component, (ii) A clustering component, (iii) And a classification component. The text-preprocessing component handles the input text's manipulation to suit the framework's solution. The clustering component is responsible for building a new dialect-aware emotion lexicon, which is later used to label each piece of text with the appropriate emotion based on the dialect it exhibits. Finally, the classification component of the framework consists of two steps: The first step is to classify the dialect used in each piece of text. The second step consists of dialect-aware classifiers that utilize the output of the first step to determine the appropriate classifier to estimate the emotion embedded in that piece of text. Figure (2) depicts the process of the framework.

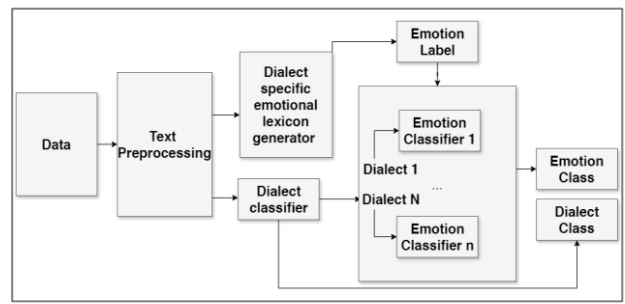

**Fig. 2:** Proposed framework

*Text Preprocessing*

This module is intended to manipulate the raw data and format it in a way that is easier for computers to process and analyze. This process is crucial for the performance of the conducted NLP task. In fact, preprocessing affects the accuracy and performance of any NLP problem. It is the first step taken for any NLP task. Some preprocessing operations include normalization and segmentation of data, tokenization of text, word stemming, and noise removal. When dealing with Arabic text, the first step is to filter out non-Arabic content from text, especially when getting content from social media. In what follows, the paper discusses the significant steps used in preprocessing in this research.

*Normalization and Segmentation*

Before tokenizing the dataset, normalization is done on Arabic diacritics such as "fatha", "damma" and "kasra" which are represented as "َ", "ُ" and "ِ" respectively. So the following word "مُذَكِراتِه" will be transformed to "مذكراته". This should help the model group similar words, albeit with a minor loss of accuracy. Arabic word segmentation separates the suffixes and prefixes attached to any word. A simple example can be seen with the word "العربية", which can be segmented to " ال+ عربي + ة " in this example. It can be seen that the prefix in this word is "ال" and the suffix is "ة" and the stemmed word is "عربي". Segmentation has been shown to significantly impact NLP applications, such as context understanding since it gives more information to the model.

*Tokenization*

The tokenization step breaks down the input text into smaller components called tokens. Several methods for performing tokenization exist, such as white-space tokenization and sub-word tokenization. White-space tokenization breaks sentences into words called tokens. This is useful for languages such as English and French, but additional steps are needed for languages such as Chinese and Japanese, where spaces do not separate words. Sub-word tokenization breaks down words into different tokens, for example, "Unfriendly" is broken down into "Un", "friend" and "ly". Tokenization has its limitations for the Arabic language, owing to the complexity of the language. Words like "عقد" and "جد", depending on the context and pronunciation, could lead to different meanings. The word "عَقَدَ" means to tie, which is different from "عَقَّدَ", which means to over-complicate. This is further complicated by using different dialects, which have their rules and vastly different sentence structure. This is also true for most languages, which is one of the tokenization's challenges.

*Clustering*

The clustering module consists of five steps that rely on two algorithms: Fast text Bojanowski *et al.* (2017) and Density-Based Spatial Clustering of Applications with Noise (DBSCAN) Ester *et al.* (1996). Fast Text transforms a given corpus into a word embedding. Word embedding allows us to measure the similarity of words in the corpus; this is a crucial step in the analysis as it allows us to study the relationship between different words. DBSCAN is a density-based clustering algorithm used to determine clusters containing words from the AEL mentioned in Section 2. A pseudocode of the DBSCAN algorithm is given in Algo. (1).

**Algorithm 1:** DBSCAN pseudo code

*Function: DBSCAN (D,eps, MinSamples)*
  *for each point k in D*
    *if k is visited*
      *continue*
    *K = visited*
    *neighborPoints = neighborQuery(k, eps)*
    *if size (neighborPoints) < MinSamples*
      *p = noise*
    *else*
      *C = new cluster*
      *updateCluster (k, neighborPoints, C, eps, MinSamples)*
*Function: updateCluster (k, neighborPoints, C, eps, MinSamples)*
  *add k to cluster C*
  *for each point p in neighborPoints*
    *if p is not visited*
      *p = visited*
      *neighborPoints' = neighborQuery (p, eps)*
      *if size(neighborPoints') >= MinSamples*
        *neighborPoints = join(neighborPoints, neighborPoints')*
    *if p is not a member of any cluster*
      *add p to cluster C*
*Function: regionQuery(k, eps)*
*return all points where their distance to k is less than eps (including k)*

The five steps are done separately on the data of each dialect as follows: (i) Transform the data into a word embedding. (ii) Find the centroid of the words for each emotion in AEL. (iii) Find the top n words outside AEL belonging to the centroid. (iv) Find words that belong to the same density-based cluster as the top n words. (v) Choose dialect-specific emotion words from the output of that cluster. Figure (3) illustrates the process of step *iii* for two emotions, where the red markers are anger emotion words from AEL the green markers are joy emotion words from AEL and the two circles represent the centroids obtained. The top n blue markers in each centroid are used as input to step iv.

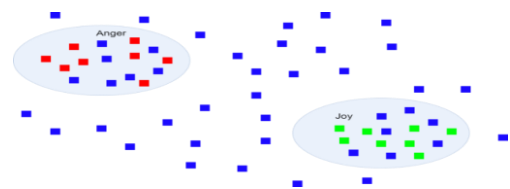

**Fig. 3:** Selection of dialect-specific emotions. Red and green markers are words from AEL

*Classification*

Transformers are used in the classification problem. Since its introduction in 2017 by the Google research team, its rapid growth has dominated the NLP field and has become a standard for any encoder/decoder model today Wolf *et al.* (2020). The transformer takes advantage of parallelization, unlike Recurrent Neural Networks (RNNs), which process data in sequential order, which is computationally more expensive Vaswani *et al.* (2017). In a high-level overview, its model architecture is divided into two major components, an encoder and a decoder. An encoder maps the input sequence to a numeric representation that holds information about the input sequence. The decoder is given the output of the encoder one sequence at a time. This allows the transformer to generate contextualized word embeddings. Contextualized word embeddings are aware of the context of words in the data. For example, the words right in "John asked Peter to do the right thing" and "Joy asked Peter to turn right at the traffic light" have different contexts. Multiple models today are built on the transformer architecture, especially in NLP. This study uses Bidirectional Encoder Representation from Transformers (BERT) Wolf *et al.* (2020). Figure (4) shows a high-level view of the BERT model.

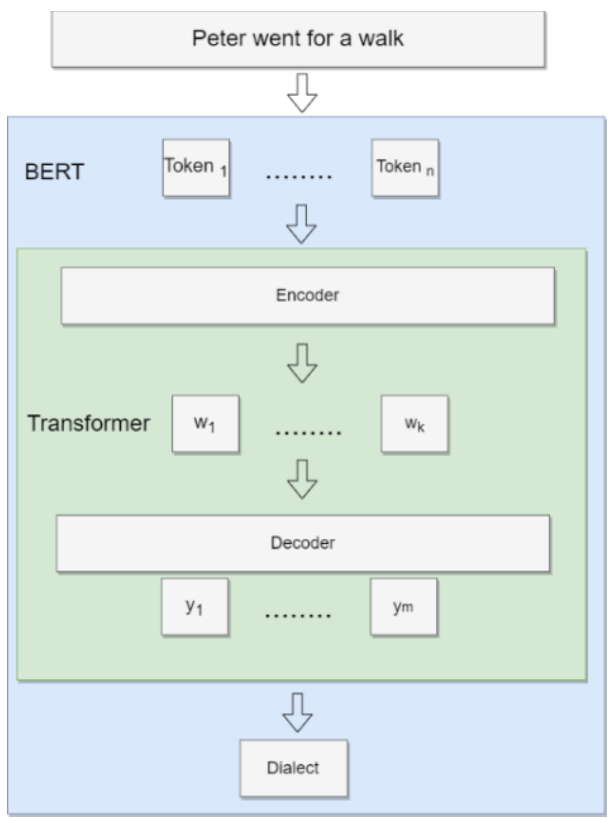

**Fig. 4:** A high-level view of BERT architecture, where n is the number of tokens, k is the length of the word embedding, and m is the length of the hidden layers

*Experiment*

In the experiments conducted in this study, the SMADC dataset is used by Alshutayri and Atwell (2019). The instances enclosed in this dataset were collected from three different sources: Facebook, Twitter, and online newspapers. For Facebook documents, they scraped 2,888,788 comments from 422,070 Facebook pages; they used Facebook Graph API to access information related to the country of origin of each page. For Twitter documents, they collected 323,236 tweets and then labeled them based on the existence of pre-defined seed words, the location of the tweet's sender, and the Geolocation of the tweet. For online newspapers, they collected 10,096 comments from 25 newspapers and automatically labeled them based on the newspaper's origin. After automatically annotating the documents, they used a novel manual annotation technique to annotate part of the dataset. They created an interactive online quiz where users log in and manually annotate several documents. Control documents were placed to check if the user was randomly choosing options and annotation conflicts were resolved by choosing majority voting resulting in 24,060 manually annotated documents. Figure (5) gives the distribution of the dialects over the three data sources.

*Experimental Setup*

An implementation of BERT that works with the Arabic language called AraBERT Antoun *et al.* (2020) is used for the classification setup and is fine-tuned empirically to solve the classification problem. Three different AraBERT models are tested in this research.

- AraBERT-base (543 MB)
- AraBERT-large (1.38 G)
- AraBERT-twitter (543 MB)

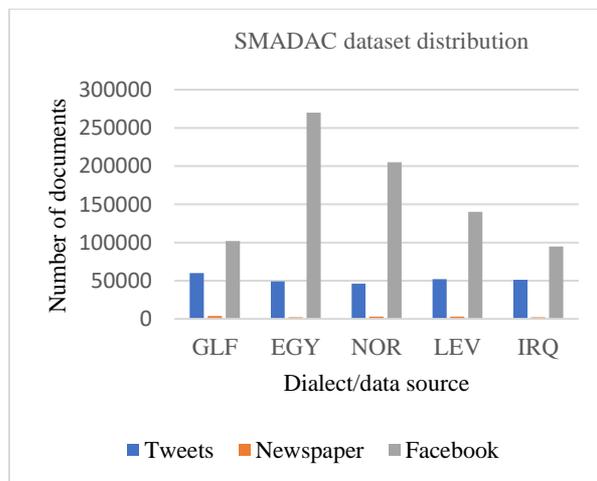

**Fig. 5:** SMADC dataset dialect distribution

Table 2: Fast text and DBSCAN parameters

| Fast text | Min count | Learning rate | Word N-Grams | Ws | Epoch |
|---|---|---|---|---|---|
| | 2 | 0.08 | 1 | 6 | 10 |
| DBSCAN | Min samples | | Eps | | Metric |
| | 9 | | 0.5 | | Manhattan |

For the clustering setup, different hyper-parameters and values were tested empirically. Table (2) reports the best-performing values for the hyper-parameters of fastText and DBSCAN. In order to avoid overfitting, the dataset was split into training/validation/testing. The training was terminated when there was no increase in performance and the results reported are for the test set.

Finally, the paper uses accuracy as the primary performance metric, defined in Eq. (1), where true positives and negatives denote documents that the model predicted correctly and false positives and negatives denote documents that the model predicted incorrectly:

$$Accuracy = \frac{True\ Positives + True\ Negatives}{True\ Positives + False\ Positives + True\ Negatives + False\ Negatives} \quad (1)$$

Precision, recall, and $f1$ score, which are given in Eqs. (2-4), respectively, are also reported. Precision tests the model's confidence in its prediction for a particular class, while recall tests the model's ability to classify all instances of a specific class correctly. The $f1$ score is the harmonized average of precision and recall:

$$Precision = \frac{True\ Positives}{True\ Positives + False\ Positives} \quad (2)$$

$$Recall = \frac{True\ Positives}{True\ Positives + False\ Negatives} \quad (3)$$

$$F1\ score = 2 * \frac{Precision * Recall}{Precision + Recall} \quad (4)$$

## Results and Discussion

The framework is tested in three ways: (i) The performance of the dialect classifier, (ii) The performance of the dialect-aware emotion classifier, and (iii) The validity of the new dialect-aware emotion lexicon. For the first point, the classifier's results are compared to other state-of-the-art techniques and the three different types of AraBERT. For the second point, the model is tested by predicting the emotion of a text using two classifiers, one that is dialect-aware and one that is not. For the third point, the validity of the results is tested by checking the new dialect-aware emotion lexicon and the newly annotated data manually for two dialects (EGY and GLF); this is due to only having access to native speakers of these dialects.

As shown in Table (3), the AraBERT model outperformed the traditional models. In addition, the most robust model is AraBERT-large having a slight edge over AraBERT-base. AraBERT-twitter that pre-trained using tweets did worse than the other AraBERT models. In addition, consistent with previous studies, MNB achieves a very high result, which is almost at par with deep learning algorithms.

Figure (6) precision results indicate that the most straightforward dialect to predict is NOR. This is not surprising since the NOR dialect differs significantly from other Arabic dialects.

The confusion matrix in Fig. (7) shows that the model can confuse GLF with IRQ and vice versa. This is expected since the two dialects are more related than the rest. An additional test for the dialect classifier was done using the AOC data and comparing the results to the state-of-the-art results obtained by Elaraby and Abdul-Mageed (2018). As shown in Table (4), the proposed framework increased the accuracy by 6.5 percentage points.

Table 3: dialect classification results

| Model | Accuracy |
|---|---|
| AraBERT-large | 89.2% |
| AraBERT-base | 87.2% |
| AraBERT-Twitter | 82.6% |
| Linear SVM | 74.7% |
| MultinomialNaiveBayes | 86.5% |
| RandomForest | 76.0% |
| Alshutari [4] | 69.2% |

Table 4: AOC prediction accuracy results on MSA, GLF, EGY, and LEV

| AraBERT-large | 88.9% |
|---|---|
| BiLSTM [10] | 82.45% |

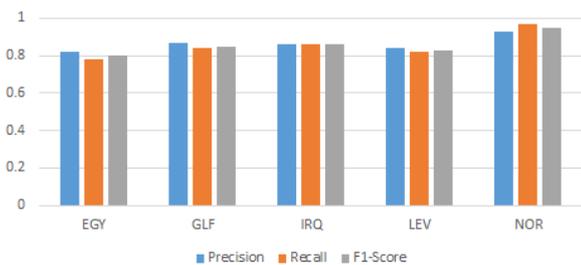

Fig. 6: Bert-large-arabertv2 performance metrics

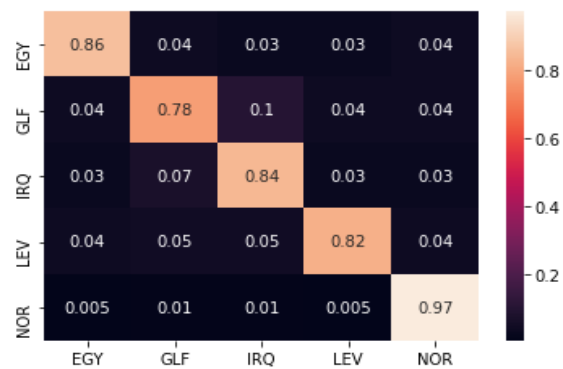

Fig. 7: Bert-large-arabertv2 confusion matrix

Table (5) reports the results for the dialect-aware emotion and general emotion classifiers. The dialect-aware emotion classifier outperforms the general classifier in both dialects by 4-5 percentage points. This indicates that emotions in different dialects exhibit different features.

Table (6) details the precision score for each emotion. The most challenging emotion to predict is "surprise" for both dialects and "sadness" for GLF, while other emotions tend to be easier to predict.

Table (7) gives a sample of the dialect-aware emotion lexicon for EGY and GLF in four emotions. Three native speakers for each dialect manually verified these results. For example, the word "اخرس", which means "shut up!" belongs to EGY_anger and another word is "عقبالك", which means "may good fortune come your way as well." belongs to GLF_joy.

**Table 5:** Dialect-aware emotion classifier vs. a general classifier

| Model | Dialect | Accuracy |
|---|---|---|
| AraBERT-large_EGY | EGY | 89.1% |
| AraBERT-large_general | EGY | 84.2% |
| AraBERT-large_GLF | GLF | 79% |
| AraBERT-large_general | GLF | 75.1% |

**Table 6:** Precision results for the dialect-aware emotion classifier

| Emotion | Precision | |
|---|---|---|
|  | GLF | EGY |
| nger | 0.88 | 0.95 |
| Disgust | 0.96 | 0.90 |
| Fear | 1 | 0.8 |
| Joy | 0.87 | 1 |
| Sadness | 0.79 | 0.87 |
| Surprise | 0.6 | 0.5 |

**Table 7:** Dialect emotions sample for EGY and GLF

| EGY_anger | GLF_anger | EGY_disgust | GLF_disgust |
|---|---|---|---|
| تخرس | كريه | تقرف | لطخ |
| اخرس | متخلف | مؤرف | بذيء |
| جلنف | مبزره | متقرفوناش | أشمط |
| بزيئ | كرهي | تقرفينا | جرذ |
| اندال | لطخ | هيقرفه | جرثوم |
| ندل | حقير | يمسخكم | وصخ |
| هكرهك | مصخره | صرصار | يغث |

| EGY_joy | GLF_Joy | EGY_sadness | GLF_sadness |
|---|---|---|---|
| هيرضى | ماقصرت | مبزعلش | بيكيني |
| تمزحو | سالخير | مخنوق | قهر |
| ميحرمناش | مشكور | اتبكي | تبكيك |
| متشكر | عقبالك | بيكو | خنق |
| هتستمتع | يهنيكم | هعيط | خذلان |
| هفرح | نكتة | مصحتش | يحزني |
| هتضحك | لهف | مضيقاني | أحزن |
| مريحني | بشوفتك | متزعلش | أبكى |
| تزغرط | ينحب | بتضايقنى | ضايقني |

## Conclusion and Future Work

The Arabic language and its dialects present considerable challenges in the field of NLP. Many topics in Arabic NLP remain an open area of research. With the recent developments in deep learning and NLP, coupled with enhanced online sources for data collection, problems such as dialect and emotion detection have received renewed interest.

This study presents a unique framework for identifying an Arabic piece of text's dialect and emotion using supervised and unsupervised models. The presented framework outperformed other models and the results obtained in previous studies. In addition, the paper presented a novel lexicon generator method used to generate an emotional lexicon for two dialects. Furthermore, the paper developed a dialect-based emotion classifier for two dialects. Given enough data, the emotion lexicon generator can be used to generate an emotion lexicon in any dialect with the supervision of a native speaker of that dialect.

Despite the positive outcomes of this study, there remains one drawback. A native speaker of each dialect must verify the output of the dialect-specific emotion lexicon generator. Such a restriction is negligible, as it only needs a native speaker and not a linguist specializing in dialect analysis of Arabic. Extensions of the current work can be added by creating emotion lexicons for additional Arabic dialects and expanding them to estimate personality traits and other mental markers for each dialect).

## Acknowledgment

This study was supported by the Research Center of the College of Computer and Information Sciences at King Saud University, Riyadh, Saudi Arabia. The author is grateful for this support.

## Funding Information

The author received no specific funding for this study.

## Ethics

The author declares that they have no ethical issues to report regarding the present study.

### Conflicts of Interest

The author declares that they have no conflicts of interest to report regarding the present study.